\documentclass[letterpaper, 10 pt, conference]{ieeeconf}  

\IEEEoverridecommandlockouts                              

\usepackage{graphics} 
\usepackage{graphicx}
\usepackage{epsfig} 
\usepackage{mathptmx} 
\usepackage{times} 
\usepackage{amsmath} 
\usepackage{amssymb}  
\usepackage{multirow}
\usepackage{xcolor}
\usepackage{bm}
\usepackage[normalem]{ulem}
\usepackage{colortbl}
\definecolor{lightgreen}{rgb}{0.80, 0.92, 0.77} 
\definecolor{lightyellow}{rgb}{1.0, 0.94, 0.76} 
\definecolor{lightblue}{rgb}{0.75, 0.85, 0.85}  
\useunder{\uline}{\ul}{}
\title{\LARGE \bf
VIPeR: Visual Incremental Place Recognition with Adaptive Mining and Continual Learning
}

\author{Yuhang Ming$^{1*}$, Minyang Xu$^{1*}$, Xingrui Yang$^{2}$, Weicai Ye$^{3}$, Weihan Wang$^{4}$, Yong Peng$^{1\S}$, \\
Weichen Dai$^{1}$ and Wanzeng Kong$^{1}$
\thanks{*Equal contribution, 
\S Corresponding author
}
\thanks{$^{1}$School of Computer Science, Hangzhou Dianzi University and Key Laboratory of Brain Machine Collaborative Intelligence of Zhejiang Province, Hangzhou, 310018, China.
        {\tt\small \{yuhang.ming, minyangxu, yongpeng, weichendai, kongwanzeng\}@hdu.edu.edu}
}%
\thanks{$^{2}$High-speed Aerodynamics Institute, CARDC, Mianyang, 621000, China.
        {\tt\small xingruiy@gmail.com}
}%
\thanks{$^{3}$State Key Lab of CAD\&CG, Zhejiang University, Hangzhou, 310058, China. {\tt\small weicaiye@zju.edu.cn}
}%
\thanks{$^{4}$Stevens Institute of Technology, Hoboken, NJ, USA, 07030, {\tt\small wwang103@stevens.edu}
}%
\thanks{This research has been supported in part by the National Natural Science Foundation of China under Grant 62401188 and Zhejiang Provincial Natural Science Foundation of China under Grant LQN25F030015.}
}

\begin{document}

\maketitle
\thispagestyle{empty}
\pagestyle{empty}

\begin{abstract}
Visual place recognition (VPR) is essential to many autonomous
systems. 
Existing VPR methods demonstrate attractive performance at the cost of 
limited generalizability. When deployed in unseen environments, these methods exhibit significant performance drops. 
Targeting this issue, we present VIPeR, a novel approach for visual incremental place recognition with the ability to adapt to new environments while retaining the performance of previous ones. 
We first introduce an adaptive mining strategy that balances the performance within a single environment and the generalizability across multiple environments. Then, to prevent catastrophic forgetting in continual learning, we 
design a novel multi-stage memory bank for explicit rehearsal.
Additionally, we propose a probabilistic knowledge distillation to explicitly safeguard the previously learned knowledge. 
We evaluate our proposed VIPeR on three large-scale datasets---Oxford Robotcar, Nordland, and TartanAir. For comparison, we first set a baseline performance with naive finetuning. Then, several more recent continual learning methods are compared. Our VIPeR achieves better performance in almost all aspects with the biggest improvement of $13.85\%$ in average performance. 

\end{abstract}

\section{Introduction}
Visual place recognition (VPR) aims to recognize a previously visited place by giving visual observations like images. It is a core component in simultaneous localization and mapping (SLAM) systems~\cite{fdslam} to ensure the robust operation of autonomous agents
in large-scale environments. 
When solving the VPR task, most of the existing approaches~\cite{netvlad, cgisnet, aegisnet} cast it as a retrieval task. The solution then typically involves a two-step process, with step one generating global descriptors from local features and step two matching the descriptors to the ones in a place-tagged database.

Traditionally, VPR methods use hand-craft local features or semantic features and construct global descriptors with vector of locally aggregated descriptors (VLAD)~\cite{vlad}, graph-based descriptors~\cite{objreloc}, \textit{etc}. More recently, following the success of NetVLAD~\cite{netvlad}, there have been notable advancements in end-to-end VPR methods in terms of accuracy~\cite{cgisnet, aegisnet} and robustness~\cite{fusionvlad, lajoie2023ral}. 
\textcolor{black}{
However, these trained models have a strong independent and identical distribution (i.i.d.) assumption on the test data, the performance drops considerably when the test data is collected from an unseen environment that doesn't follow the i.i.d. assumption.
}


\begin{figure}[t]
    \centering
    \includegraphics[width=0.95\linewidth]{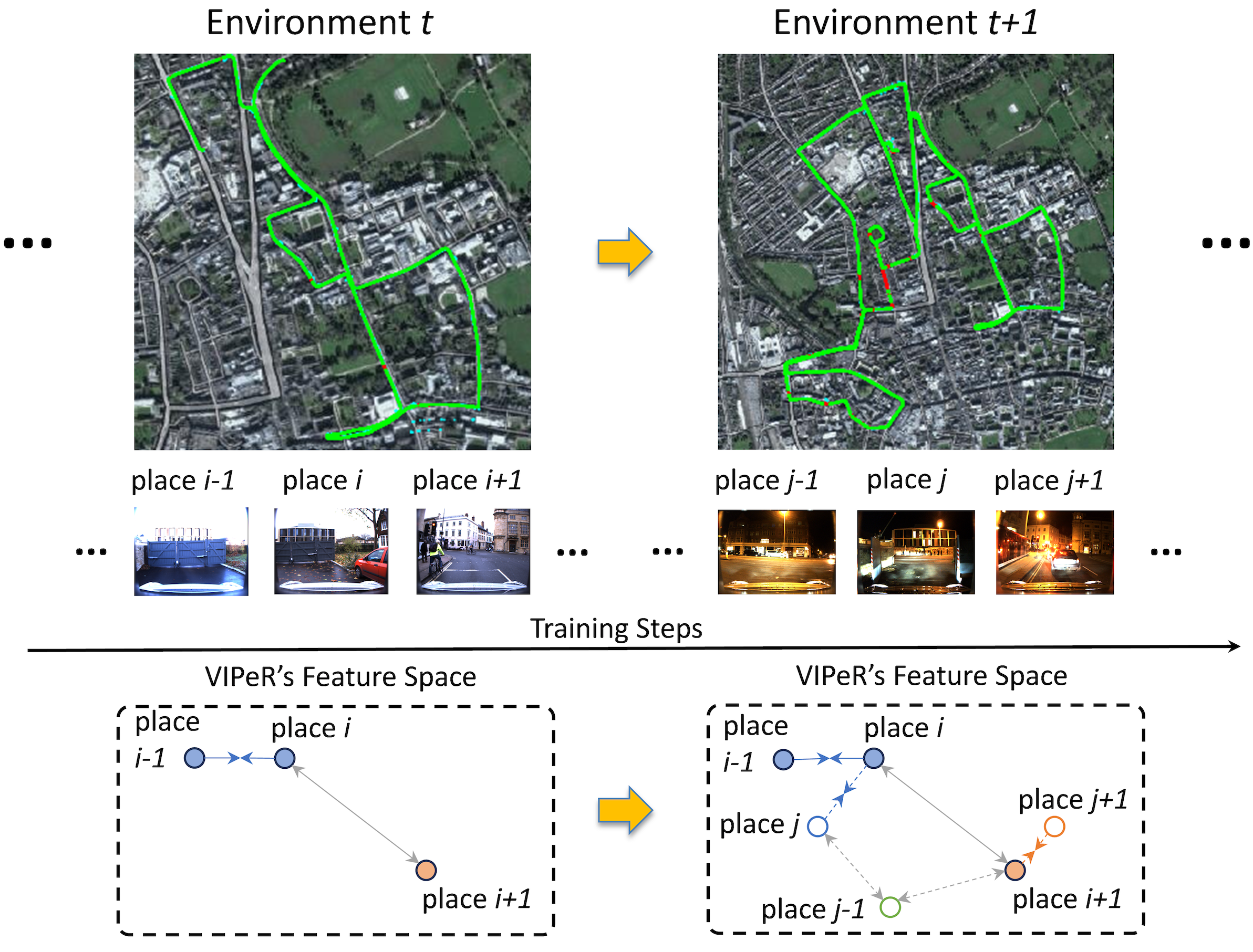}
\vspace{-2ex}
    \caption{
    Visual Incremental Place Recognition (VIPeR).
    Within a single environment, VIPeR aims to bring descriptors from the same place closer together and push those from different places farther apart in a learned feature space. 
    When adapting to multiple environments, VIPeR is geared towards acquiring knowledge about new places while retaining information about previously encountered ones.
    }
    \label{fig:viper}
\vspace{-5ex}
\end{figure}

However, when deploying the VPR-equipped autonomous agents and AR/VR systems in real-world scenarios, it is difficult to collect sufficient data that covers all potential environments beforehand to pre-train the deep learning-based models. Such a gap between the advance in deep learning-based VPR methods and the practical needs of autonomous agents and AR/VR systems leads to the fact that recent practical-oriented systems still prefer the traditional VPR methods~\cite{vial2024jofr, egolocate}. In this paper, we aim to close this gap by enabling the continual learning ability of the deep learning-based VPR method. In particular, we present VIPeR, a visual incremental place recognition method that can adapt to the newly observed environment during deployment while reserving its knowledge of previously visited environments. Fig.~\ref{fig:viper} illustrates the core idea of the proposed VIPeR.

The idea of enabling continual learning for deep neural networks is not new. Also known as continual learning or incremental learning, it is designed to overcome the notorious catastrophic forgetting when learning from sequential inputs. Despite that continual learning has been widely studied~\cite{cl-survey} and applied to many computer vision~\cite{cl-classification} or natural language processing~\cite{cl-nlp} tasks, there is only a handful of works attempting to apply it to the place recognition task. 
AirLoop~\cite{airloop} and InCloud~\cite{incloud}, as the pioneers in continual place recognition, both tackle the catastrophic forgetting with rehearsal-based and regularization-based methods.

Although achieving promising performance, there is still plenty of room for improvement. As pointed out in CCL~\cite{ccl}, the naive triplet loss used in AirLoop~\cite{airloop} and InCloud~\cite{incloud} inevitably introduces bias when learning the global representation of various places. Such a phenomenon becomes even severe when hard-mining is applied to the training. In addition, the naive choice of a simple memory structure substantially hinders the power of the rehearsal strategy. Finally, the deterministic knowledge distillation used in AirLoop and InCloud is very sensitive to the outliers in the data, leading to potential performance degradation. 

Targeting these issues, we first introduce an adaptive mining scheme for triplet loss in metric learning. Our adaptive mining scheme can adaptively choose hard mining or soft mining, achieving a balanced performance between a single environment and multiple environments.
Focusing on continual learning, 
\textcolor{black}{our memory bank differs from ~\cite{a-gem,bioslam}, achieves excellent results without any additional computational burden.}
Our memory bank is made of three components. For the current environment, we have a sensory memory bank and a working memory bank. For previously visited environments, we have another long-term memory bank.
Finally, to make the model more robust against outliers, we extend the deterministic knowledge distillation used in AirLoop~\cite{airloop} and InCloud~\cite{incloud} to a probabilistic one.
We conduct extensive experiments on three large-scale datasets and our proposed VIPeR demonstrates superior performance when compared to other state-of-the-art (SoTA) continual place recognition methods.

In summary, our contributions are as follows: 
\begin{itemize}
    \item We introduce an adaptive mining scheme for metric learning to balance the performance within a single environment and 
    across multiple environments;
    \item We design a novel, \textcolor{black}{multi-stage memory bank} for rehearsal, alleviating the effects of catastrophic forgetting;
    \item We propose a probabilistic knowledge distillation to regularize the training across different environments;
\end{itemize}

\section{related work}

\subsection{Continual Learning}

Continual learning targets the sequential learning setting and focuses on improving the adaptability and generalizability of the deep learning models. However, naive solutions like finetuning on the new tasks make the model quickly forget the knowledge about previous tasks, resulting in catastrophic forgetting~\cite{cf-dnns}. Addressing this issue, most existing methods can be categorized into rehearsal-based approaches and regularization-based approaches~\cite{cl-classification}.

The core of the rehearsal-based approach is a memory buffer that either explicitly or implicitly stores samples from previous tasks. Then, when the model is trained on a new task, the samples stored in the buffer join the samples of the current task to refresh the model's memory about previous tasks. Exemplary works in this approach are iCaRL~\cite{icarl}, gradient episodic memory (GEM)~\cite{rehearsalgem}, and deep generative replay (DGR)~\cite{dgr}, which tackle the image classification task with iCaRL and GEM explicitly stores a small number of samples per class for rehearsal and DGR accompanies an additional generative model to enable implicit rehearsal. 
\textcolor{black}{Building upon GEM, A-GEM~\cite{a-gem} reduces the computational and memory costs by averaging the gradient of a random subset of episodic memory examples. 
Our memory bank, on the other hand, favors explicit memory storage and avoids introducing any additional computational overhead.}

Shifting the focus to the model itself, the regularization-based approach applies regularization on either the weight or the function. Regarding weight regularization, it commonly evaluates the importance of the model's parameters on previous tasks and mitigates the performance degradation by introducing a penalty for the loss. Within this scope, elastic weight consolidation (EWC) ~\cite{ewc} employs the Fisher information matrix to calculate the importance, while syntactic intelligence (SI) ~\cite{si}  and memory aware synapses (MAS)~\cite{MASforget} both favor the online importance estimation. Then, in function regularization, various knowledge distillation approaches have been explored to distill the knowledge of old models to the new ones. An example work is learning without forgetting (LwF)~\cite{lwf}, which performs distillation by comparing the predictions from the output head of the old tasks to the ones of new tasks.



\subsection{Learning-based Place Recognition}
\textcolor{black}{
NetVLAD~\cite{netvlad}, a milestone in learning-based VPR methods, introduced differentiable VLAD~\cite{vlad} descriptors, enabling end-to-end training for VPR tasks. Building on this, SPE-VLAD~\cite{spevlad} incorporated a spatial pyramid structure to encode structural information within the VLAD descriptor. Extending these advancements to 3D inputs, various methods employ 3D backbones, such as FusionVLAD~\cite{fusionvlad}, which focuses on cross-viewpoint VPR, and AEGIS-Net~\cite{aegisnet}, which emphasizes aggregating multi-level features.
}

\textcolor{black}{
In addition to NetVLAD~\cite{netvlad}, 
other pooling-based methods have also been explored for generating global descriptors, owing to their simplicity and generalizability.
Examples include sum pooling in APANet~\cite{apanet}, max pooling by Gordo \textit{et al.}~\cite{gordo2017ijcv}, and generalized-mean pooling (GeM)~\cite{gem} adopted in MinkLoc3D~\cite{minkloc3d}.
}
More recently, with Transformer~\cite{transformer} prevailing in a wide range of tasks, it has also been used to create global descriptors in VPR, \textcolor{black}{like $R^2$Former~\cite{r2former}}. 

\textcolor{black}{
To broaden the scope of this review, we acknowledge works like CosPlace~\cite{cosplace} and EigenPlaces~\cite{eigenplaces}, which approach VPR as a classification problem. However, this paper focuses on VPR formulated as a retrieval problem.
}

\textcolor{black}{With the advancement of visual foundation models, methods based on these pre-trained models have demonstrated compelling performance. 
Among them, CricaVPR~\cite{cricavpr} and SelaVPR~\cite{selavpr} utilize pre-trained DINOv2~\cite{dinov2} and employ parameter-efficient transfer learning to adapt the model for VPR tasks. 
Focusing on global aggregation, SALAD~\cite{salad} proposes a soft-assignment of DINOv2 features to NetVLAD's clusters while SegVLAD~\cite{segvlad} augments NetVLAD descriptor with supersegments from SAM~\cite{sam}.
Finally, AnyLoc~\cite{anyloc} and VDNA-PR~\cite{vdna-pr} both explore the data representation power of the visual foundation model. In particular, VDNA-PR combines the features extracted by DINOv2 with the visual distribution of neuron activations representation, but AnyLoc experiments with a wide range of data, from indoor to outdoor, aerial to underwater.
Although the use of visual foundation models reduces the likelihood of encountering out-of-distribution data, continual learning is still indispensable, as we demonstrate in Table~\ref{tab:dino-cross}.}

Despite tremendous advances in VPR research, most of the work tackles the task under the batch learning setting with the complete dataset available prior to the training. However, our paper investigates the VPR task under a sequential learning setting, which is addressed by a few works.
More related to our work, BioSLAM~\cite{bioslam} tackles the VPR task from a continual learning perspective and chooses the generative replay from a rehearsal-based approach with a dual memory zone to store the previous observations. Nevertheless, AirLoop~\cite{airloop} combines a rehearsal-based approach with a regularization-based approach to mitigate the catastrophic forgetting issue with a lightweight model. Following this idea, InCloud~\cite{incloud} and CCL~\cite{ccl} extend to point cloud input with InCould introducing an angular-based knowledge distillation to relax the function regularization and CCL favoring contrast learning with InfoNCE~\cite{infonce} loss over metric learning with triplet loss~\cite{triplet}. 
Our VIPeR significantly differs from the above-mentioned methods with the adaptive mining for metric learning with triplet loss, a delicately designed multi-stage memory bank for rehearsal, and probabilistic knowledge distillation (PKD) for regularization.

\begin{figure*}[t]
  \centering
  \includegraphics[width=\textwidth]{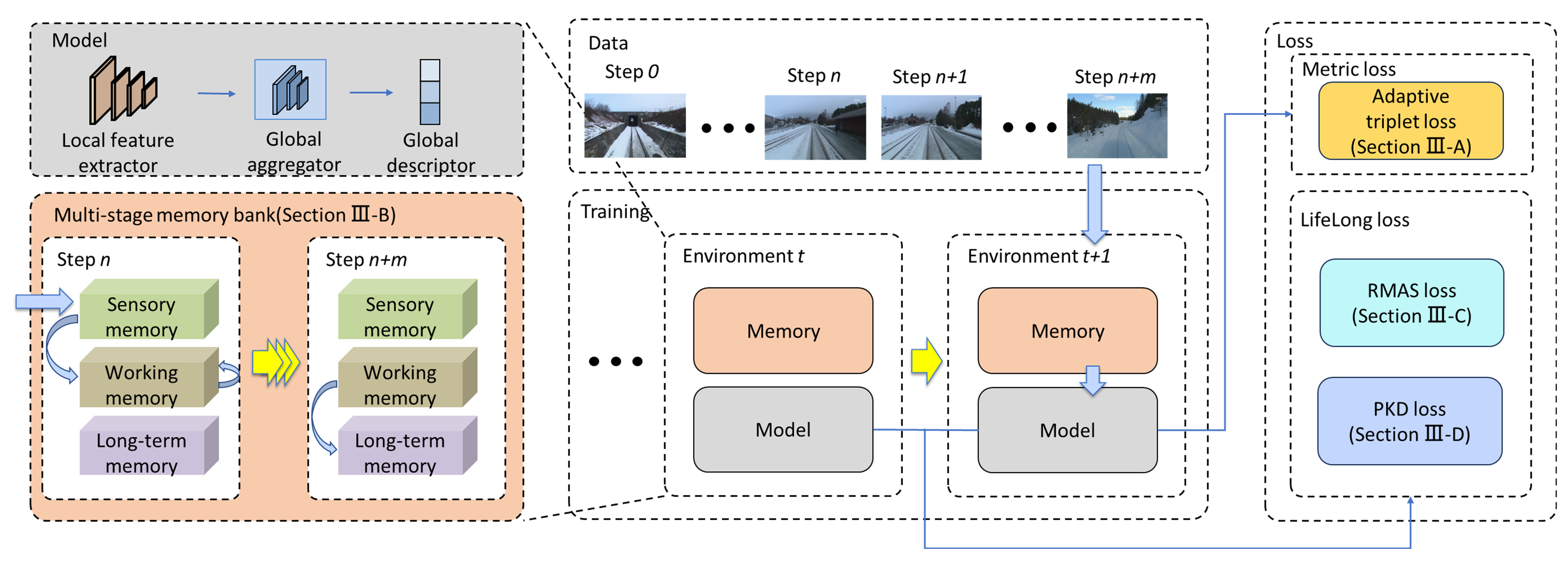}
\vspace{-6ex}
  \caption{
  Overview of main components and data flow within our proposed VIPeR. To alleviate the catastrophic forgetting in continual learning, our proposed VIPeR accompanies the place recognition model with a \textcolor{black}{multi-stage memory bank} for rehearsal, and adaptive mining, relational memory aware synapses (RMAS) and probabilistic knowledge distillation (PKD) for regularization. 
  }
  \label{fig:overivew}
\vspace{-4ex}
\end{figure*}

\section{method}
\label{sec:method}

As with many previous works on VPR, we cast the task of visual incremental place recognition as a retrieval problem. 
In particular, we define a sequence of environments $\mathcal{E} = \{E_0, E_1, ..., E_t, ...\}$ with only one environment being available for training at a time. Furthermore, we define a set of images as observations for each environment $E_t = \{I_0, I_1, ..., I_i, ...\}$, which is also acquired in a sequential manner. 
Then, the VPR task is solved by incrementally constructing a database of place-tagged global descriptors $\mathcal{G} = \{\bm{g}_0, ..., \bm{g}_i, ... \}$ with each one generated with $\bm{g}_i = f(I_i)$. 
$f(\cdot)$ here is a trainable model that consists of two sub-models, one extracts local features from input images and the other one aggregates the local features into global descriptors.
Whenever the query image $I_q$ becomes available, the same model is used to generate a global descriptor from the query image $\bm{d}_q = f(I_q)$. Following this, by matching the query descriptor to the database descriptors, we can retrieve the top-$K$ closest neighbors of the query image as the recognized places.
In the following, we present VIPeR to tackle visual incremental place recognition by combining metric learning with continual learning. An overview of VIPeR is shown in Fig.~\ref{fig:overivew}.

\subsection{Metric Learning}
\label{sec:adaptive}

To train our model, we adopt metric learning with triplet loss. 
We first construct input triplet $\mathcal{T}$ with an anchor image $I^{anc}$ that represents a place in the environment, $m$ positive images $I^{pos}=\{I^{pos}_0, I^{pos}_1, ..., I^{pos}_m\}$ that are taken from the same place, and $n$ negative images $I^{neg}=\{I^{neg}_0, I^{neg}_1, ..., I^{neg}_n\}$ that are taken from different places.
Then, given the triplet $\mathcal{T} = (I^{anc}, I^{pos}, I^{neg})$, we drew inspiration from the naive triplet loss~\cite{triplet} and hard triplet loss~\cite{hardmining}, and introduced adaptive triplet loss to train our model.

The naive triplet loss~\cite{triplet} is defined as $L_{triplet}=\max \left( s^{an}_j-s^{ap}_i+\delta ,0 \right)$,
where $s^{an}_j=sim(f(I^{anc}),f(I^{neg}_j))$ is the cosine similarity between the global descriptors generated from the anchor image and a negative image sample, $s^{ap}_i$ is the cosine similarity between the ones from the anchor and a positive image sample, $\delta$ is a hyper-parameter indicating the margin. When selecting the positive and negative image, the naive triplet loss opted for a random mining strategy with the sample indices $i$ and $j$ randomly selected. Despite that, the naive triplet loss exhibits better generalizability and is used in previous works like AirLoop~\cite{airloop} and InCloud~\cite{incloud}; it also makes the training process less effective, resulting in less discriminative global descriptors.  

The hard triplet loss~\cite{hardmining}, to the contrary, goes to another extreme and favors the hard mining strategy. When choosing the sample images, the hard triplet loss focuses on the most difficult sample with the sample indices selected as $i=\text{argmin}_i(s^{ap}_i)$ and $j=\text{argmax}_j(s^{an}_j$). Such a strategy largely increases the discrimination of the global descriptor and has been widely used~\cite{aegisnet, pointnetvlad} for place recognition in a single large-scale environment. 
However, such a mining strategy inevitably introduces bias into the training, leading to performance degradation in continual learning settings. We later demonstrate this in Table~\ref{tab:ablation-study}.

\textcolor{black}{
To balance between the hard and random mining, we introduce adaptive mining.
Unlike existing adaptive mining methods~\cite{adaptivetriplet,smart-mining} that only adjust negative samples, our method dynamically choose different positive and negative samples.
Especially, we use the fluctuation of the loss as the criterion to determine the difficulty:
\begin{equation}
    \Delta L_{triplet}=\lVert s_{j}^{an,k}-s_{i}^{ap,k}-s_{j}^{an,k-1}+s_{i}^{ap,k+1} \rVert _1
\end{equation}
where $k$ denotes the data at the $k$-th step. When the loss increment is greater than $T_d$, we consider the current training difficulty to be too high, and continued training may lead to insufficient generalization performance of the model. Therefore, with $s_{j}^{an}$ and $s_{i}^{ap}$ sorted in descending and ascending orders, we decrease $i$ and increase $j$ to reduce the training difficulty. Contrarily, when the loss decrement is greater than $T_e$, we increase $i$ and decrease $j$ to increase the training difficulty. By adjusting the training difficulty through $\Delta L_{triplet}$, both positive and negative samples can be considered, ensuring the model's generalization ability.}

\subsection{\textcolor{black}{Multi-stage Memory Bank}}
\label{sec:memory}

To alleviate the notorious catastrophic forgetting in continual learning, we first investigate the rehearsal-based approach to refresh the model's memory about previous environments. Specifically, we present a \textcolor{black}{multi-stage memory bank} $\mathcal{B}=(M, R)$ with a fixed-size image storage space $M$ and a corresponding adjacency matrix $R$.
Each element in the adjacency matrix $R_{i,j}$ represents the relative location between place $i$ and place $j$, which is then used to determine whether these two places belong to a positive or a negative pair.

Regarding the fixed-size image storage space $M$, we mimic the human memory system and design the architecture as a composition of a per-environment sensory memory $M^{sn}$, a per-environment working memory $M^{wk}$, and an environment-agnostic long-term memory $M^{lt}$.
Our \textcolor{black}{multi-stage memory bank} is capable of improving the model's adaptability in a new environment while preserving its stability over all previous environments. We illustrate the update process of the proposed \textcolor{black}{multi-stage memory bank} in Fig.~\ref{fig:overivew}. 

For any unseen environment $t$, the sensory memory $M^{sn}_t$, which is a first-in-first-out queue of size $l^{sn}$, is always first activated and used to store the most recently visited places. With such a design, we ensure the model stays sensitive to the latest changes in the scene. 
Whenever the sensory memory is full, the working memory $M^{wk}_t$, as a list of size $l^{wk}$, is activated to store the places coming out of the sensory memory. When the working memory is also full, we decide on whether the new place should be stored with a probability $p=l^{wk} / num$, where $num$ is the total number of images that have passed in the current environment. This makes the working memory a perfect complement to the sensory memory since the probability of updating the working memory keeps decreasing as more images are processed, allowing the old observations to be preserved.

The long-term memory \begin{math}M^{lt}\end{math}, also as a list of size $l^{lt}$, stores the samples from all previously visited environments to ensure its environment-agnostic property. 
When entering a new environment $t$, the long-term memory $M_t^{lt}$ is updated with $M_{t}^{lt}=\omega M_{t-1}^{wk}+\left( 1-\omega \right) M_{t-1}^{lt}$,
where \begin{math}M^{wk}_{t-1}\end{math} and \begin{math}M^{lt}_{t-1}\end{math} are the working memory and the long-term of environment $t-1$, \begin{math}\omega\end{math} is a hyper-parameter to control the update ratio.

  

\subsection{Relational Memory Aware Synapses}
Additionally, we explore a weight regularization approach to prevent undesired weight updates when the model is adapting to the new environment.
Particularly, we follow the relational memory aware synapses (RMAS) proposed in AirLoop~\cite{airloop} to estimate the importance weight $\Omega^{RMAS}_{t}$ of each parameter $\theta$ in the environment $t$. Then, the changes in the parameters are penalized with a regularization loss $L_{RMAS}$. Note that for simplicity, we omit the subscript $t$ to index the environment in the loss function. We hereby briefly review the formulation of RMAS. For more details, we refer the readers to the original AirLoop~\cite{airloop} paper. 

For environment $t$, the weight is approximated as
\vspace{-1ex}
\begin{equation}
    \Omega^{RMAS}_{t} \approx \frac{1}{N_t}\sum_{k=1}^{N_t} \left( \frac{\partial \lVert \tilde{S}_{k, t} \rVert_F}{\partial \theta} \right) ^2,
\vspace{-1ex}
\end{equation}

where $N_t$ is the number of images that have been processed, $\lVert \cdot \rVert_F$ is the Frobenius norm,  $\tilde{S}_{k, t}\in \mathbb{R}^{3\times3}$ is the Gram matrix at the $k$-th training step. Each element in the Gram matrix is computed using the triplets we selected with adaptive mining in Section~\ref{sec:adaptive}, following $\tilde{S}_{k, t|(0,0)} = s^{aa}$, $\tilde{S}_{k, t|(0,1)} = s^{ap}_i$, $\tilde{S}_{k, t|(0,2)} = s^{an}_j$, \textit{etc.}.

Then, the RMAS loss is computed as 
\vspace{-1ex}
\begin{equation}
    L_{RMAS}=\sum_{\theta \in \Theta}^n \Omega^{RMAS}_{t-1} \left( \theta_t - \theta_{t-1} \right) ^2 ,
\vspace{-1ex}
\end{equation}
where $\Theta$ is the set of all parameters, $\theta_t$ and $\theta_{t-1}$ indicate the updated parameter in the current environment $t$ and the frozen parameter after training on environments $1, 2, ..., t-1$.

\subsection{Probabilistic Knowledge Distillation}
Moreover, we delve into the function regularization approach with knowledge distillation to further enhance the model's adaptability to unseen environments during continual learning. 
Although previous works 
have also explored knowledge distillation, we believe that their performance is limited due to two reasons. First of all, when selecting memory samples for distillation, CCL~\cite{ccl} use images from the past environment. We believe that using data unseen by the old model for knowledge distillation helps to extract the common knowledge, thereby enhancing the model's generalization ability. 
Secondly, these works like AirLoop~\cite{airloop} and InCloud~\cite{incloud} opt for deterministic distillation with angular metrics, which makes the model more focused on local structures of samples.  

To select more effective samples, we propose to use the samples from our \textcolor{black}{multi-stage memory bank} $M$ to perform knowledge distillation.
Besides, we present PKD to make the model pay more attention to the global distribution of the samples. To estimate the distribution of the samples in the learned feature space, we first construct the matrices $H_t$ and $H_{t-1}$ using the model trained in the current environment $t$ and the frozen model that has been trained on environment $1, 2, ..., t-1$ respectively. Specifically, elements in the matrix are computed as
\vspace{-2ex}
\begin{equation}
    H_{\Xi|i,j}=\frac{\bm{g}_{i, \Xi} \cdot \bm{g}_{j, \Xi}^T}{\sqrt{d}} \;\;\; \text{for} \;\;\; \Xi \in [t-1, t] ,
\vspace{-1ex}
\end{equation}
where $d$ is the dimension of the global descriptor $\bm{g}$. Although similar ideas can be seen in CCL~\cite{ccl}, we notice that their design with the distillation temperature hyper-parameter easily leads to a vanishing gradient during the training process. We are inspired by Transformer~\cite{transformer} and make a simple yet effective modification by replacing the distillation temperature hyper-parameter with a model-specific value $\sqrt{d}$ to prevent the gradients vanishing issues.
\begin{figure*}
    \centering
    \includegraphics[width=1\linewidth]{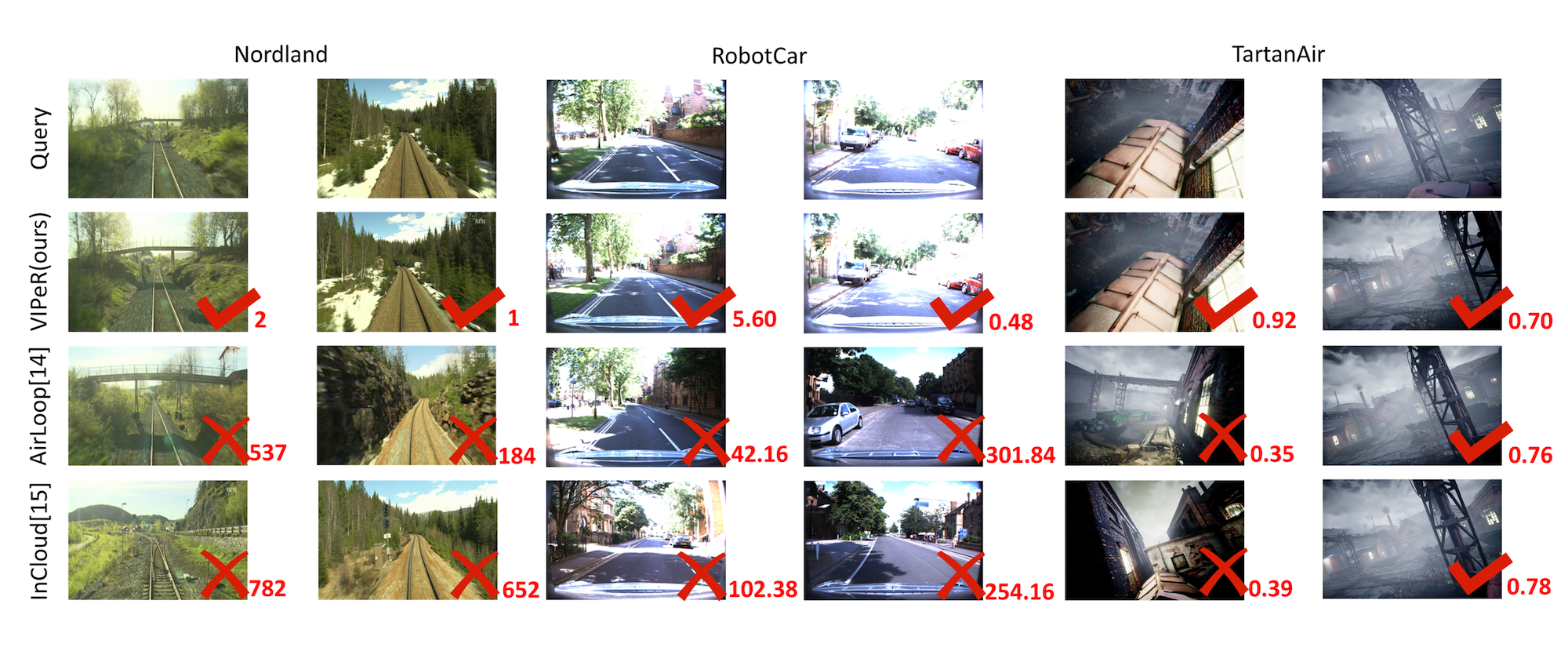}
\vspace{-8ex}
    \caption{\textbf{Top-$1$ retrieval} of the model trained for $T$ environments and evaluated on the first environment. Our VIPeR can better discriminate similar places and exhibits better resilience to catastrophic forgetting. \textcolor{black}{The numbers on the image indicate the distance from the query image; the \textbf{timestamp distance$\downarrow$} is measured in Nordland, the \textbf{Euclidean distance$\downarrow$} in RobotCar, and the \textbf{sIoU$\uparrow$} in TartanAir.}}
    \label{fig:retrieval}
\vspace{-4ex}
\end{figure*}
Once the $H$ matrices are computed, we can estimate the distribution with the softmax (SM) function and calculate the PKD loss with Kullback-Leibler (KL) divergence as 
\begin{equation}
    L_{PKD}=\sum_{i,j}{KL\left\{ \text{SM} \left( H_{t-1|i,j} \right) \log \left(\text{SM} \left( H_{t|i,j} \right) \right) \right\}} .
\end{equation}

\subsection{Final Loss}
Finally, we combine the above-mentioned loss functions into the final loss function of the following form
\vspace{-1ex}
\begin{equation}
    L=L_{ada-triplet} + \lambda_1 L_{RMAS} + \lambda_2 L_{PKD},
    \label{eq:combined}
\end{equation}
where \begin{math}\lambda_1\end{math} and \begin{math}\lambda_2\end{math} are hyper-parameters used to adjust the contributions of the RMAS and PKD loss functions.

\section{experiment}
\subsection{Implementation Details}
When choosing the VPR method, we first experiment with the same VPR method used in AirLoop~\cite{airloop}, which consists of a pre-trained VGG-19~\cite{vgg} for local feature extraction and GeM~\cite{gem} for global descriptor aggregation.
The hyper-parameter in GeM also follows AirLoop as $p=3$, and the dimension of its global descriptor is set to $d=1024$. Additionally, to get more discriminative global descriptors, we extend the VPR method by replacing the GeM layer with a NetVLAD layer~\cite{netvlad}. We especially follow many other VPR methods like~\cite{netvlad, pointnetvlad, cgisnet, aegisnet} and set the number of clusters to $K=64$. However, we set the dimension of global descriptors generated by NetVLAD to be the same as that of GeM descriptors.

When forming the triplet tuple, we empirically set the number of positive and negative samples to be $m=1$ and $n=5$, and we set the margin to be $\delta=1$.
Regarding adaptive mining, we prevent the model from always learning with easy samples by explicitly setting the threshold for using easier samples to be higher than the one for using more difficult ones.
Especially, we choose \textcolor{black}{$T_d=0.05$} and $T_e=0.03$.

As for the \textcolor{black}{multi-stage memory bank}, 
we ensure a fair comparison by setting the total size of our memory bank to be the same as the queue's length in AirLoop~\cite{airloop}. Particularly, we empirically divide the total size into $l^{sn}=500$ for sensory memory, $l^{wk}=400$ for working memory, and $l^{lt}=100$ for long-term memory. In addition, we set the hyper-parameter in the memory bank to be $\omega=0.5$. 

Finally, we train the VPR methods using the stochastic gradient descent (SGD) optimizer with a learning rate of $0.002$ and a momentum of $0.9$.
\textcolor{black}{All experiments in this paper are conducted on a single NVIDIA RTX 4090 GPU and each dataset was trained for 2 to 3 hours.}


\subsection{Datasets and Evaluation Metrics}

\textbf{Datasets:} We evaluate our proposed VIPeR on three large-scale, publicly available datasets, namely Oxford RobotCar~\cite{robotcar}, Nordland~\cite{nordland}, and TartanAir~\cite{tartanair}.

Oxford RobotCar~\cite{robotcar} is a real-world autonomous driving dataset with data collected under various weather conditions across multiple urban environments in Oxford. 
For evaluation, we follow AirLoop~\cite{airloop} and select three environments under the tag ``sun'', ``overcast'', and ``night''. For each environment, the same two sequences are selected for training and testing. 
We define two images belonging to the same place if the distance between them is less than 10m and the yaw difference is less than $15^\circ$.

Nordland~\cite{nordland} is also a real-world dataset with across-season data collected from a railway journey through the Nordland region of Norway. 
As AirLoop~\cite{airloop}, we also use the recommended train-test split and determine whether two images are taken from the same place if they are separated by three images or less.

TartanAir~\cite{tartanair} is a photo-realistic synthetic dataset. It covers both indoor and outdoor environments and provides data collected under diverse lighting conditions, weather conditions, and times of the day.
Also following AirLoop~\cite{airloop}, we select 5 environments for evaluation with the same surface intersection-over-union (sIoU) used as the criteria to decide whether two images are from the same place. Specifically, we set sIoU$<0.1$ for negative, sIoU$>0.7$, and sIoU$>0.5$ for positive in training and test, respectively.


\textbf{Evaluation metrics:} In line with AirLoop \cite{airloop}, we use the recall rate at 100\% precision as the core metric to evaluate the model's performance on single environments. 
To better assess the performance in continual learning, we evaluate the model trained after each environment on every environment in the environment sequence $\mathcal{E}$ of length $T$, even the unseen ones. We summarize the results in an evaluation matrix $P \in \mathbb{R}^{T\times T}$ with each of its element $P_{i,j}$ being the recall rate at 100\% precision when evaluating the model trained after environment $i$ in environment $j$.

Given the evaluation matrix, we further compute the average performance (AP), backward transfer (BWT), and forward transfer (FWT), which are introduced in~\cite{rehearsalgem}, to evaluate the overall performance, how does learning in new environments affect the previous knowledge, and how well does the model generalize to unseen environments. We refer readers to~\cite{rehearsalgem} for the detailed formulas for computation.


\subsection{Evaluation and Discussions}
We first set a baseline performance with naive finetuning on both of the VPR methods we experimented with. In addition, we also explore the performance of two generic weight regularization methods in continual learning, EWC~\cite{ewc} and SI~\cite{si}. For more recent methods, we compare our proposed VIPeR to AirLoop~\cite{airloop} and InCloud~\cite{incloud}, which are specifically designed for visual incremental place recognition. 
Since AirLoop only supports the combination of VGG-19~\cite{vgg} and GeM~\cite{gem}, and InCloud takes in only LiDAR point clouds, both are modified and re-trained, using their official implementation, to get comparison results.
\textcolor{black}{Despite our eager to compare with BioSLAM~\cite{bioslam}, neither of their code or data were publicly available. 
}

We present the qualitative results in Fig.~\ref{fig:retrieval} with VIPeR, AirLoop~\cite{airloop}, and InCloud~\cite{incloud} trained for all available environments and evaluated on the very first one to demonstrate the model's resilience to catastrophic forgetting. 
Then, we also present quantitative results in all three environments in Table~\ref{tab:performance}.  
It can be seen that naive finetuning and generic weight regularization method EWC~\cite{ewc} and SI~\cite{si} exhibit inferior performance, especially in AP and BWT, indicating their inadequate abilities when fighting against catastrophic forgetting in visual incremental place recognition.
Furthermore, despite that both AirLoop \cite{airloop} and InCloud \cite{incloud} achieved improved performance over baseline methods by utilizing a combination of several continual learning methods to mitigate the catastrophic forgetting, they suffer from a notable performance drop when changing from GeM~\cite{gem} to more discriminative NetVLAD~\cite{netvlad}.

The results of our proposed VIPeR exhibit superior performance over other methods in almost all metrics.
In particular, our VIPeR prevails in terms of AP and FWT across all three datasets, which implies the model's strong performance and good generalizability towards unseen environments.
In terms of BWT, which measures the forgetness of the model, our VIPeR didn't to achieve the best performance. However, we argue this amount forgetting is acceptable as even after forgetting some knowledge of the previous visited place, our VIPeR still exhibits better recognition performance when compared to AirLoop and InCloud.
Additionally, when changing the global descriptor aggregator from GeM to NetVLAD, our VIPeR, unlike AirLoop~\cite{airloop} or InCloud~\cite{incloud}, achieves even better performance, which further demonstrates the effectiveness of presented solutions.

\textcolor{black}{
We further highlight the importance of continual learning within the context of foundation models, a necessity that becomes evident when examining models like AnyLoc~\cite{anyloc}, a DINOv2-based place recognition model with broad compatibility across diverse data types. To ensure a fair comparison, we utilize AnyLoc pre-trained specifically for urban and indoor environments and adjust our backbone to DINOv2~\cite{dinov2}.
We challenge our VIPeR by conducting experiments not only on single datasets but also across multiple datasets—TartanAir, Nordland, and RobotCar—testing the model’s robustness under varied conditions. 
Table~\ref{tab:dino-cross} presents our quantitative results, where VIPeR outperforms AnyLoc by a substantial margin across all three datasets. Notably, when using the VGG~\cite{vgg} backbone, VIPeR’s cross-dataset training slightly diminishes performance. However, with DINOv2 as the backbone, cross-dataset performance not only matches but surpasses single-dataset results, demonstrating enhanced effectiveness of continual learning in conjunction with visual foundation models for robust place recognition.
}

\begin{table*}[t]
\centering
\caption{Performance of continual place recognition methods. 
The ones marked with $^\S$ are directly taken from AirLoop~\cite{airloop}.
We mark the \textbf{best results} and the \ul{second best}. }
\label{tab:performance}
\vspace{-2ex}
\scalebox{1}{
\begin{tabular}{l|l|l|ccccccccc}
\hline
\multicolumn{1}{c|}{\multirow{2}{*}{Backbone}} & \multicolumn{1}{c|}{\multirow{2}{*}{\begin{tabular}[c]{@{}c@{}}Feature\\ Aggregator\end{tabular}}} & \multirow{2}{*}{Method} & \multicolumn{3}{c}{Oxford RobotCar} & \multicolumn{3}{c}{Nordland} & \multicolumn{3}{c}{TartanAir} \\ \cline{4-12} 

 &  & \multicolumn{1}{c|}{} & AP$\uparrow$ & BWT$\uparrow$ & FWT$\uparrow$ & AP$\uparrow$ & BWT$\uparrow$ & FWT$\uparrow$ & AP$\uparrow$ & BWT$\uparrow$ & FWT$\uparrow$ \\ \hline
\multirow{11}{*}{VGG~\cite{vgg}} & \multirow{8}{*}{GeM~\cite{gem}} 
& $^\S$Finetune & 0.411 & -0.066 & 0.462 & 0.615 & -0.012 & 0.549 & 0.754 & -0.009 & 0.730 \\
 &  & $^\S$EWC~\cite{ewc} & 0.416 & -0.054 & 0.461 & 0.614 & -0.014 & 0.549 & 0.758 & -0.005 & 0.728 \\
 &  & $^\S$SI~\cite{si} & 0.407 & -0.062 & 0.454 & 0.614 & -0.010 & 0.549 & 0.753 & -0.010 & 0.730 \\
 &  & $^\S$AirLoop~\cite{airloop} & 0.461 & -0.013 & 0.485 & 0.631 & 0.018 & 0.546 & 0.769 & 0.007 & 0.736 \\
 &  & InCloud~\cite{incloud} & {\ul 0.491} & -0.003 & {\ul 0.507} & 0.624 & 0.017 & 0.532 & 0.776 & {\ul 0.018} & 0.749 \\
 &  & VIPeR  & 0.490 & -0.030 & 0.503 & {\ul 0.639} & 0.020 & {\ul 0.554} & 0.752 & -0.005 & 0.741 \\ \cline{2-12} 
 & \multirow{4}{*}{NetVLAD~\cite{netvlad}} & Finetune & 0.379 & -0.068 & 0.413 & 0.641 & -0.006 & 0.543 & 0.770 & 0.009 & {\ul 0.759} \\
 &  & AirLoop~\cite{airloop} & 0.416 & -0.065 & 0.501 & {\ul 0.639} & -0.001 & 0.549 & {\ul 0.782} & \textbf{0.021} & 0.756 \\
 &  & InCloud~\cite{incloud} & 0.445 & \textbf{0.050} & 0.413 & 0.622 & \textbf{0.039} & 0.505 & 0.771 & {\ul 0.018} & 0.754 \\
 &  & VIPeR & \textbf{0.559} & {\ul 0.001} &  \textbf{0.569} & \textbf{0.704} & {\ul 0.021} & \textbf{0.628} & \textbf{0.800} & 0.011 &  \textbf{0.779} \\ \hline
\end{tabular}
}
\vspace{-4ex}
\end{table*}

\begin{table}[t]
\centering
\caption{\textcolor{black}{Performance in AP$\uparrow$ with different backbones and training setups. We mark the \textbf{best results} and the \ul{second best}. }}
\label{tab:dino-cross}
\scalebox{0.9}{
\vspace{-2ex}
\begin{tabular}{l|l|ccc}
\hline
Model
& 
Method
& TartanAir & Nordland & RobotCar \\
\hline

\multirow{4}{1.7cm}{DINOv2~\cite{dinov2} + NetVLAD~\cite{netvlad}} 
 & AnyLoc-urban~\cite{anyloc} & 0.678 & 0.543 & 0.647 \\
 & AnyLoc-indoor~\cite{anyloc} & 0.679 & 0.483 & 0.602 \\
 & VIPeR (S.D.M.S)$^1$ & \textbf{0.807} & {\ul 0.709} & {\ul 0.677} \\
 & VIPeR (M.D.M.S)$^2$ & \textbf{0.807} & \textbf{0.711} & \textbf{0.680} \\
 \hline
 \multirow{2}{1.7cm}{VGG~\cite{vgg} + NetVLAD~\cite{netvlad}} & VIPeR (S.D.M.S)$^1$ & {\ul 0.800} & 0.704 & 0.559 \\
 & VIPeR (M.D.M.S)$^2$ & {\ul 0.800} & 0.663 & 0.528 \\ \hline
 \multicolumn{5}{l}{%
  \begin{minipage}{\linewidth}%
    \footnotesize 
    $^1$ S.D.M.S. stands for single-dataset, multiple-scene, means a model is trained for one dataset with different scenes;\\
    $^2$ M.D.M.S. is multiple-dataset, multiple-scene, means the model is trained across all three datasets in the order of TartanAir, Nordland, and RobotCar.
  \end{minipage}%
}\\
\end{tabular}
}
\vspace{-3ex}
\end{table}

\subsection{Ablation Studies}

\begin{figure}[t]
    \centering
    \scalebox{1}{
    \includegraphics[width=\linewidth]{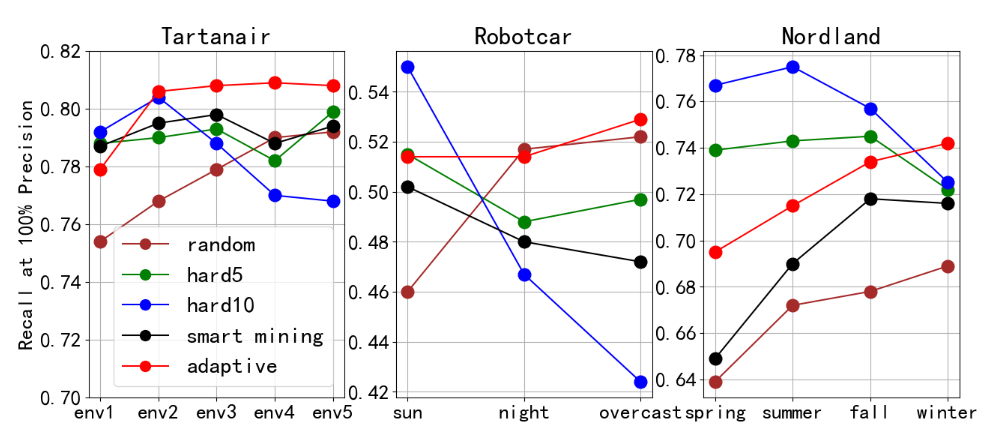}
    }
    \vspace{-5ex}
    \caption{\textcolor{black}{Performance evaluated in the first environment after training in all environments with different mining strategies, highlighting forgetfulness. ``hard5'', ``hard10'' indicate the performance of using the hard mining strategy with 5, and 10 negative samples.}}
    \label{fig:mining}
    \vspace{-4ex}
\end{figure}

\begin{table}[t]
\centering
\caption{\textcolor{black}{Ablation study with results in AP$\uparrow$. We mark the \textbf{best results} and the \ul{second best}. }}
\scalebox{0.95}{
\begin{tabular}{l|ccc}

\hline
Method & TartanAir & Nordland & RobotCar \\ \hline
VIPeR (default) & \textbf{0.800} & {\ul 0.704} &  \textbf{0.559}\\ 
\hline
VIPeR (random mining) & 0.784 & 0.649 & 0.547 \\
VIPeR (5 hard mining) & 0.786 & 0.702 & 0.537 \\
VIPeR (10 hard mining) & 0.778 & \textbf{0.725} & 0.495 \\
VIPeR (smart mining) & 0.789 & 0.681 & 0.514 \\ \hline
VIPeR (naive queue) & 0.758 & 0.632 & 0.460 \\ \hline
VIPeR (RKD) & 0.793 & 0.698 & 0.520 \\
VIPeR (AKD) & {\ul 0.796} & 0.683 & {\ul 0.554} \\ \hline
\end{tabular}
\label{tab:ablation-study}
}
\vspace{-4ex}
\end{table}


\textbf{Adaptive Mining:} 
We first validate the contribution of our adaptive mining by comparing to the performance using random mining, hard mining, and smart mining~\cite{smart-mining}.
With the results from the three datasets shown in Fig.~\ref{fig:mining} and Table~\ref{tab:ablation-study},
we can see that the adaptive mining strategy proposed in our VIPeR exhibits favorable performances, especially in TartanAir~\cite{tartanair} and Oxford RobotCar~\cite{robotcar}.
Although our adaptive mining does not perform as good as hard mining with 10 negative samples in Nordland~\cite{nordland},
we believe it is mainly caused by the less divergent scenes in Nordland, reducing the impact of the bias. 

\textbf{Multi-stage Memory Bank:} Then, we replace the multi-stage memory bank with the naive queue used in AirLoop~\cite{airloop} and InCloud~\cite{incloud},
and present the quantitative results in Table ~\ref{tab:ablation-study}. It is notable that the \textcolor{black}{multi-stage memory bank} surpasses the naive queue by a large margin across all datasets, demonstrating the superior performance of the \textcolor{black}{multi-stage memory bank} in continual learning with good generalizability and resilience to catastrophic forgetting.

\textbf{Probabilistic Knowledge Distillation:} Finally, the contribution of PKD is evalutated by replacing with relational knowledge distillation (RKD) from AirLoop~\cite{airloop} and angular-based knowledge distillation (AKD) from InCloud~\cite{incloud}.
As shown in Table~\ref{tab:ablation-study}, it is evident that our PKD exceeds the performance of RKD and AKD, validating that our distillation approach is capable of grasping the underlying data distribution, which is a big advantage in dealing with catastrophic forgetting and the adaptability when training in unseen environments.

\vspace{-1ex}
\section{Conclusions}
In this work, we present VIPeR, a visual incremental place recognition method. 
We dive into triplet mining strategies and find a balance between per-environment and cross-environment performance with adaptive mining. 
In addition, 
we design a novel \textcolor{black}{multi-stage memory bank} and present a probabilistic knowledge distillation to mitigate the catastrophic forgetting.
Through extensive experiments on large-scale datasets, we demonstrate the superior of the our VIPeR over not only other visual increment place recognition methods, but also visual foundation model-based methods.
As for future work, we believe there are many possible directions, like extending the VIPeR model to handle other forms of single-modality input or cross-modality input.





\bibliographystyle{ieeetr}
\bibliography{root}

\end{document}